%% file: main.tex
\documentclass[10pt,twocolumn,letterpaper]{article}

\usepackage[T1]{fontenc}
\usepackage[margin=0.75in,columnsep=0.25in]{geometry}
\usepackage[hyphens]{url}
\usepackage{graphicx}
\usepackage{natbib}
\usepackage{caption}
\usepackage{amsmath,amssymb,amsfonts,amsthm,mathtools}
\usepackage{booktabs}
\usepackage{nicefrac}
\usepackage{microtype}
\usepackage{subcaption}
\usepackage{multirow}
\usepackage{array}

\usepackage[colorlinks=true,linkcolor=blue,citecolor=blue,urlcolor=blue]{hyperref}
\hypersetup{pdftitle={HiPACE: Hierarchical Phase-Boundary Analysis and Controlled Evaluation of Feature Absorption in Sparse Autoencoders},pdfauthor={Jinyuan Zhang, Peng He, Yin Yuan, He Hu, ShengShuo Jiao}}
\usepackage[capitalize,noabbrev]{cleveref}

\theoremstyle{plain}
\newtheorem{theorem}{Theorem}
\newtheorem{proposition}[theorem]{Proposition}

\theoremstyle{definition}

\theoremstyle{remark}

\crefname{assumption}{Assumption}{Assumptions}
\Crefname{assumption}{Assumption}{Assumptions}

\input{notation}

\begin{document}

\twocolumn[{%
  \begin{center}
    {\Large\bfseries HiPACE: Hierarchical Phase-Boundary Analysis and Controlled Evaluation\\
     of Feature Absorption in Sparse Autoencoders\par}
    \vspace{1.3em}
    {\normalsize
    \begin{tabular}{c@{\hspace{4em}}c}
      Jinyuan Zhang & Peng He$^{\ast}$\\
      {\ttfamily 202621116012480@stu.hubu.edu.cn} & {\ttfamily penghe@hubu.edu.cn}\\
      \addlinespace[3pt]
      Yin Yuan & He Hu\\
      {\ttfamily 202521120012766@stu.hubu.edu.cn} & {\ttfamily 202521120012751@stu.hubu.edu.cn}\\
      \addlinespace[3pt]
      ShengShuo Jiao & \\
      {\ttfamily 202621120012764@stu.hubu.edu.cn} & \\
    \end{tabular}\par}
    \vspace{1.1em}
    {\itshape Hubei University, Wuhan, China\par}
    \vspace{0.5em}
    {$^{\ast}$Corresponding author: {\ttfamily penghe@hubu.edu.cn}\par}
    \vspace{1.2em}
    \begin{minipage}{0.92\textwidth}
      \begin{abstract}
\input{sections/0_abstract}
      \end{abstract}
    \end{minipage}
  \end{center}
  \vspace{2.2em}
}]

\input{sections/1_introduction}
\input{sections/2_related_work}
\input{sections/3_theory}
\input{sections/4_evaluation}
\input{sections/5_experiments}
\input{sections/7_discussion}

\bibliographystyle{plainnat}
\bibliography{references}

\appendix
\input{sections/A_appendix}
\input{sections/B_preregistered_proxy}

\end{document}

%% file: notation.tex
\newcommand{\Rn}{\mathbb{R}}

\newcommand{\lc}{\lambda_c}
\newcommand{\absmag}{\mathrm{abs\_mag}}

%% file: sections/0_abstract.tex
Sparse autoencoders (SAEs) decompose LLM activations into sparse dictionary
atoms, aiming to give each distinct concept its own feature. One recurring
behavior complicates this premise: \emph{feature absorption}, in which a
parent concept and its children---\emph{fruit} and \{\emph{apple},
\emph{banana}, \emph{pear}\}, say---collapse into a shared family direction
rather than remain distinct features. Prior work documents absorption
empirically; what is missing is a closed-form prediction of when the shared
direction is the cost-optimal representation of an active semantic family.
This paper closes that gap. For a hierarchical Bernoulli generator with $k$
active children and residual scale $\alpha$, the $L_0$-penalized
reconstruction objective admits a closed-form phase boundary
$\lambda_c(k,\alpha)=\alpha^2 k/(k-1)$: above it, pure parent absorption is
strictly cheaper than pure child coding. Building on this boundary, we
introduce \textsc{HiPACE}, an evaluation protocol that tests the boundary's
structural consequence in real SAE dictionaries---measuring parent--child
decoder structure over WordNet families, freezing the discovery-selected
statistic before testing on unseen families, and contrasting genuine
families against randomized sibling nulls. The boundary proves sharp in its
native regime: it predicts the synthetic transition within $\pm15\%$ on all
30 tested cells. In Pythia-160m SAEs, the parent--child decoder gap recovers
the predicted ordering with partial correlations of $-0.681$ at layer 6 and
$-0.933$ at layer 11; the locked holdout sustains the effect at $-0.605$ and
$-0.914$, outside the sibling nulls ($p=0.002$, the floor for 500 draws).
Controlled activation composition then connects the theory's active-child
count to the recovered family directions, and residual-stream interventions
show that signed family directions increase parent-category logits,
reversing under sign flip and vanishing under random controls. Together with
comparable child-direction effects, this establishes causal sufficiency at
the family-subspace level---the granularity at which the theory makes its
prediction.

%% file: sections/1_introduction.tex
\section{Introduction}
\label{sec:intro}

Sparse autoencoders (SAEs) decompose LLM activations into sparse combinations
of dictionary atoms and are widely used in post-hoc mechanistic
interpretability \citep{bricken2023monosemanticity,cunningham2023sparse,gao2024scaling}.
A recurring failure mode is \emph{feature absorption}: when a representation
contains a parent concept and several children---\emph{fruit} and
\{\emph{apple}, \emph{banana}, \emph{pear}\}, for example---the learned
dictionary may allocate capacity to a shared family direction rather than to
distinct child features. Existing work documents this behavior empirically
\citep{chanin2024absorption}, but does not predict when absorption is the
cost-optimal representation.

We study absorption as competition between two coding regimes. A compact
regime represents an active semantic family through its shared parent
direction; a distributed regime preserves the child-specific residuals. For
a hierarchical Bernoulli generator, comparing their reconstruction and
sparsity costs yields a closed-form transition at
$\lambda_c(k,\alpha)=\alpha^2 k/(k-1)$, where $k$ is the number of active
children and $\alpha$ is the child-residual scale. This perspective turns
absorption from a purely descriptive observation into a testable phase
boundary.

To connect this idealized prediction to learned dictionaries, we introduce
\textsc{HiPACE} (\textbf{Hi}erarchical \textbf{P}hase-Boundary
\textbf{A}nalysis and \textbf{C}ontrolled \textbf{E}valuation). The empirical
challenge is matching the measurement to the object predicted
by the theory. The boundary concerns how representation is allocated across a
semantic family, so we evaluate parent--child structure in the learned SAE
dictionary rather than requiring a particular latent to win on each prompt.
Our evaluation separates metric selection on discovery families from a frozen
test on unseen WordNet families, and uses randomized sibling groups to test
whether recovery depends on genuine hierarchy. We then connect the idealized
active-count variable to real activations and test the behavioral relevance of
the recovered family directions through residual-stream intervention.

\begin{figure*}[t]
  \centering
  \includegraphics[width=0.84\textwidth]{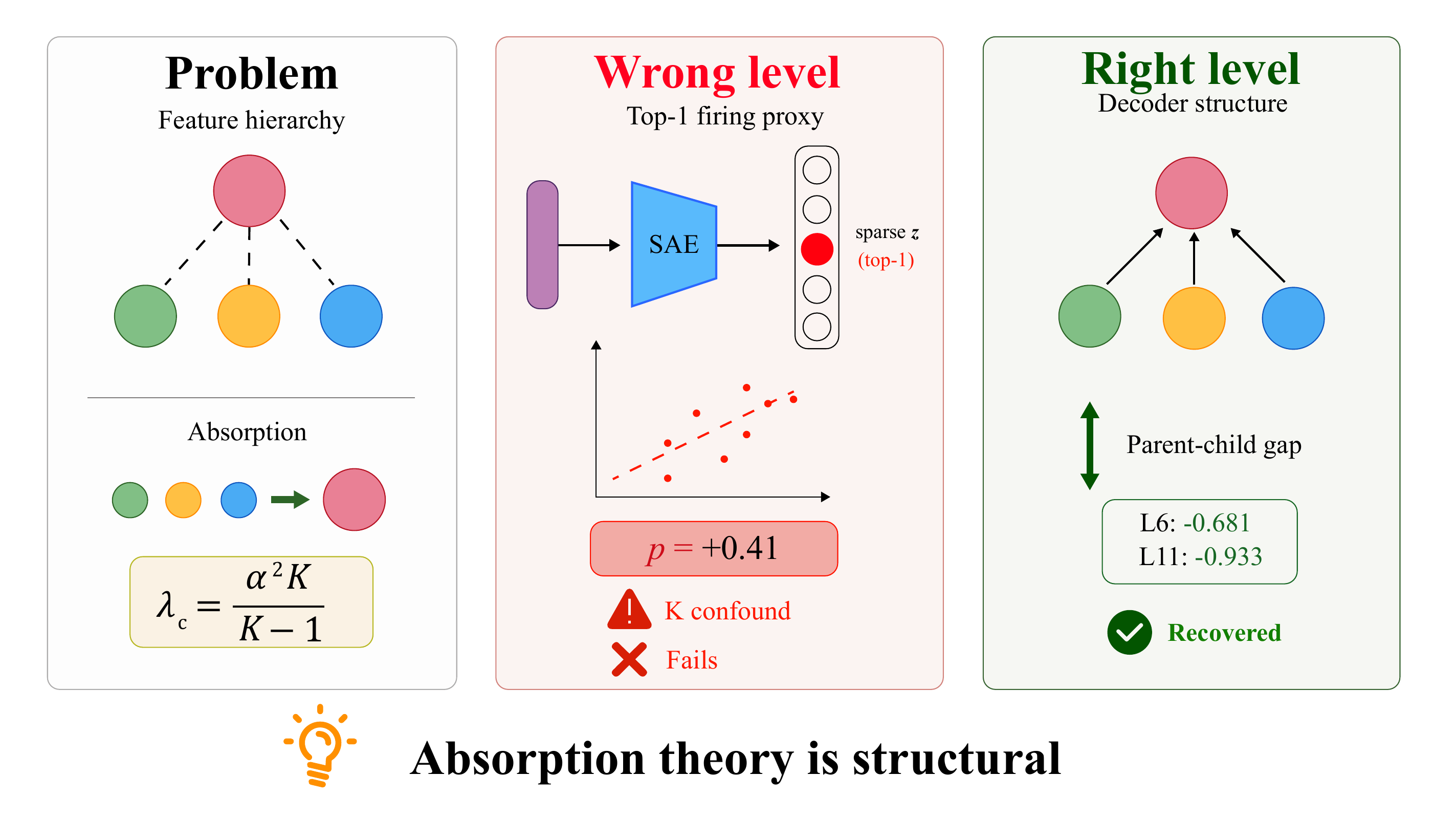}
  \caption{\textbf{Motivation for family-level evaluation.}
  Hierarchical feature competition predicts a phase boundary in dictionary
  structure. The empirical analysis therefore measures parent--child
  structure over semantic families, validates the selected statistic on a
  frozen holdout, and tests the recovered family directions mechanistically.
  Here $k$ denotes active-child count, whereas $K$ denotes family size.}
  \label{fig:motivation-overview}
\end{figure*}

We make three contributions:
\begin{enumerate}
  \item \textbf{A phase-boundary theory for hierarchical coding.} We derive
  the active-count threshold, characterize mixed dictionaries, and state
  conditional shifts for Matryoshka and Archetypal SAEs.
  \item \textbf{The HiPACE family-level evaluation.} The synthetic boundary
  matches all 30 tested cells. In real Pythia-160m SAEs, parent--child decoder
  structure recovers the predicted ordering on discovery families and on a
  frozen WordNet holdout, with randomized sibling groups as a specificity
  control.
  \item \textbf{Mechanistic evidence with explicit scope.} Controlled
  activation composition recovers the predicted active-count response, and
  signed L6 family-direction interventions change parent-category logits.
  Parent and child directions have comparable effects, localizing the causal
  effect to the family subspace rather than to a unique parent atom.
\end{enumerate}

\paragraph{Organization.}
\Cref{sec:related} situates the work; \Cref{sec:methodology} presents HiPACE;
\Cref{sec:experiments} states the research questions and answers them
experimentally; and \Cref{sec:discussion} discusses implications and scope.

%% file: sections/2_related_work.tex
\section{Related Work}
\label{sec:related}

\paragraph{Sparse autoencoders and dictionary design.}
SAEs decompose LLM activations into sparse combinations of dictionary atoms
and are a standard tool for post-hoc interpretability
\citep{bricken2023monosemanticity,cunningham2023sparse}. Variants differ in
how sparsity is enforced and in what structure the dictionary is pushed
toward. TopK SAEs fix the active-set size directly and improve the
reconstruction--sparsity frontier \citep{gao2024scaling}; JumpReLU SAEs
replace the $L_1$ penalty with thresholded gating
\citep{rajamanoharan2024jumprelu}. Matryoshka SAEs train nested
sub-dictionaries so that outer atoms capture the most universal directions
\citep{bussmann2025matryoshka}, while Archetypal SAEs constrain decoder
atoms to convex combinations of training samples \citep{fel2025archetypal}.
Public releases such as Gemma Scope \citep{lieberum2024gemmascope} turn these
design choices into auditable dictionaries. Each design reshapes the cost
landscape under which child features compete with a shared parent direction,
and our theory treats two of them explicitly: Propositions~\ref{prop:matry}
and~\ref{prop:arch}
predict downward threshold shifts for Matryoshka and Archetypal training, and
the synthetic sweeps confirm both shifts (\S\ref{sec:synthetic}).

\paragraph{Feature absorption and hedging.}
\citet{chanin2024absorption} introduced absorption as the systematic failure
of SAE latents to fire on arbitrary subsets of their declared concept---a
``starts with S'' latent skipping particular words---and showed the pathology
worsens with width. Feature hedging \citep{chanin2025hedging} describes a
complementary failure: the encoder direction that reads into a latent
mismatches the decoder direction that writes it back. Both lines are
empirical. They document \emph{that} latents collapse and propose
architectural fixes, but neither predicts \emph{when} collapse becomes the
cost-optimal representation. HiPACE closes that gap with a closed-form
threshold and tests of its structural consequences. Hedging does not account
for our structural results: the encoder/decoder cosine of Pythia-160m modal
latents is $+0.94$ at the median (\S\ref{sec:experiments}).

\paragraph{Phase transitions in sparse coding.}
Threshold analysis has a long history in sparse coding. Classical
formulations \citep{olshausen1996emergence} and k-sparse autoencoders
\citep{makhzani2013ksparse} set the optimization backdrop;
compressed-sensing theory predicts sharp phase transitions in $L_1$ recovery
\citep{donoho2009message}, and Toy Models of Superposition
\citep{elhage2022toymodels} established the sharp-threshold lens for features
in superposition. These results are asymptotic, or concern flat feature
structure in toy generators. Our derivation specializes the lens to
\emph{hierarchical} features---a parent direction plus child residuals---and
yields the critical penalty in closed form rather than as an asymptotic
constant. Beyond verifying the threshold in its synthetic regime, we test its
structural consequence in learned dictionaries.

\paragraph{Identifiability and evaluation.}
Whether a uniquely correct SAE dictionary exists is open: independently
trained SAEs recover inconsistent decompositions of the same activations
\citep{leask2025canonical}, while a single SAE can be trained across many
models at once \citep{thasarathan2025universal}. Benchmarks therefore measure
what learned dictionaries do. SAEBench \citep{karvonen2025saebench}
consolidates instance- and dictionary-level tests (TPP, SCR, RAVEL,
absorption, sparse probing) and finds Matryoshka variants strongest at
moderate $L_0$. Such evaluations score latents one prompt or one feature at
a time. The absorption boundary instead predicts how representation is
allocated across a semantic family, so our evaluation audits parent--child
decoder structure at the family level---a generator-side prediction that does
not depend on a unique recovered dictionary---and freezes the statistic on
discovery families before scoring unseen WordNet families.

\paragraph{Positioning.}
Taken together, prior work documents absorption empirically, varies the
architecture that produces it, analyzes thresholds in classical or toy
regimes, and benchmarks dictionaries feature by feature. This paper differs
on three axes: a closed-form phase boundary for hierarchical absorption; a
family-level structural audit matched to the theory's unit of prediction; and
a frozen-holdout protocol with randomized sibling nulls as the confirmatory
test.

%% file: sections/3_theory.tex
\section{Methodology}
\label{sec:methodology}

\begin{figure*}[t]
  \centering
  \includegraphics[width=0.98\textwidth]{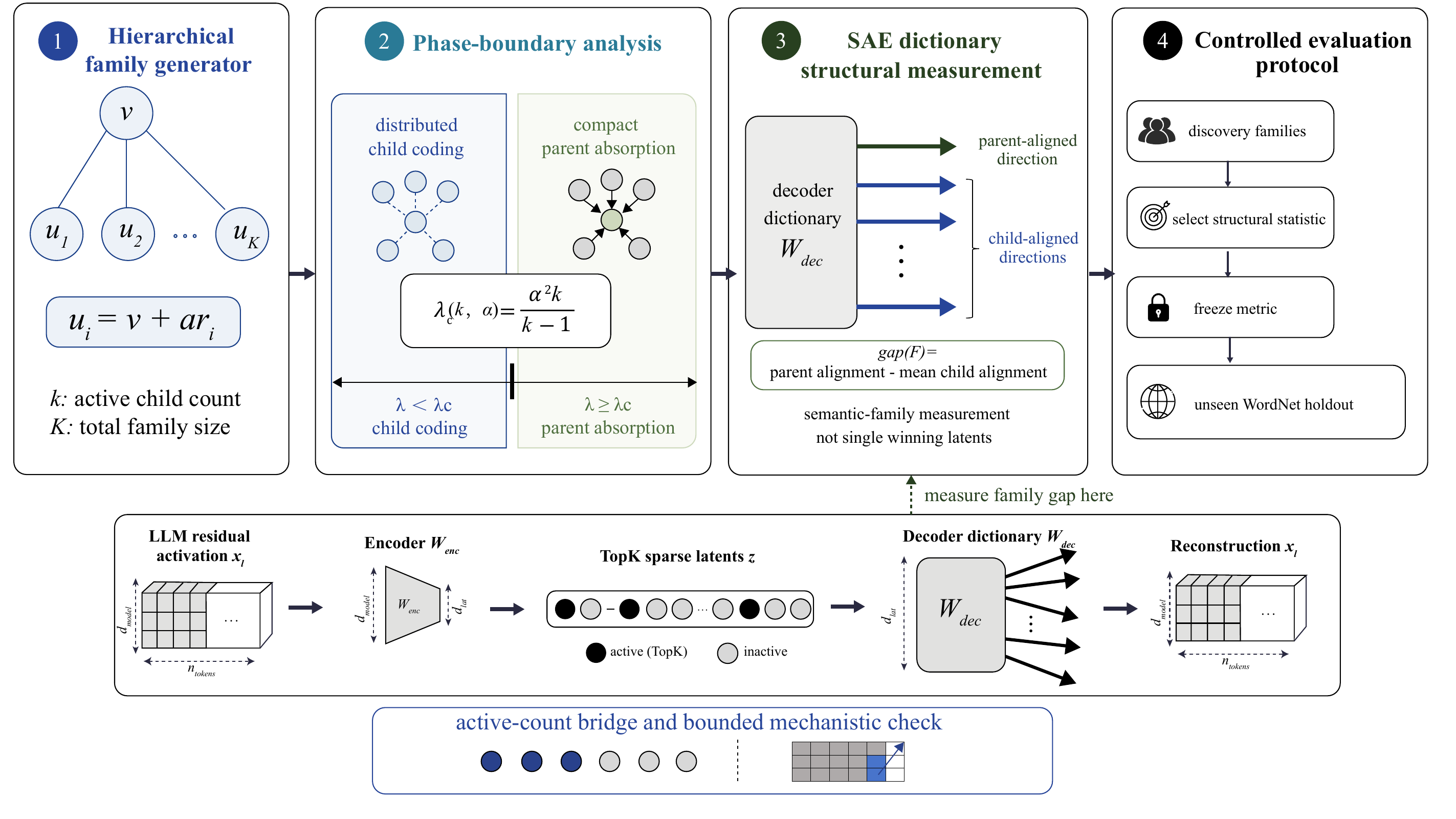}
  \caption{\textbf{Overview of HiPACE.} The framework connects a hierarchical
  family model and its phase-boundary analysis to family-level SAE dictionary
  measurement, frozen validation on unseen semantic families, and controlled
  mechanistic follow-up. Lowercase $k$ denotes sample-level active-child count;
  uppercase $K$ denotes total family size.}
  \label{fig:hipace-structure}
\end{figure*}

\paragraph{Framework overview.}
As summarized in \Cref{fig:hipace-structure}, HiPACE has four components.
The first is a hierarchical family model that separates a shared parent
direction from child residuals. A phase-boundary analysis then identifies
the sparsity level at which compact parent coding becomes cheaper than
distributed child coding. For real SAE dictionaries, HiPACE translates this
prediction into a family-level parent--child structural measurement, and the
evaluation design separates discovery from frozen holdout validation, with
controlled activation composition and residual-stream intervention as
mechanistic follow-up. We define these components before presenting the
research questions and experimental settings in \Cref{sec:experiments}.

\subsection{Hierarchical Phase-Boundary Analysis}
\label{sec:theory}

\paragraph{Generative model.}
Let $v\in\Rn^d$ be a unit-norm parent direction and let $r_1,\dots,r_K$ be
unit-norm residual directions, mutually orthogonal and each orthogonal to
$v$:
\begin{align}
  \langle v,r_i\rangle &= 0, \label{eq:orth-v}\\
  \langle r_i,r_j\rangle &= \delta_{ij}. \label{eq:orth-r}
\end{align}
Define $K$ \emph{child atoms} by
\begin{equation}
u_i \;=\; v\;+\;\alpha\,r_i,\qquad i=1,\dots,K,
\label{eq:children}
\end{equation}
where $\alpha>0$ controls per-child residual magnitude. Each child is
independently active with probability $p$; with active set $S\subseteq[K]$
of size $k=\lvert S\rvert$, the activation is
\begin{equation}
  x=\sum_{i\in S}u_i
   =k\,v+\alpha\sum_{i\in S}r_i .
\label{eq:hbg-activation}
\end{equation}
We call this the
hierarchical Bernoulli generator (HBG).

\paragraph{Sparse-coding objective.}
We work with the $L_0$-penalized reconstruction objective
\begin{equation}
\mathcal{J}_\lambda(z) \;=\; \lVert x - Dz\rVert^2 + \lambda\,\lVert z\rVert_0,
\label{eq:obj}
\end{equation}
for a dictionary $D$ and sparse code $z$. TopK SAEs instead impose a hard
firing budget, which we use as an empirical analogue of the penalized
objective.\footnote{The multiplier $\lambda$ prices sparsity marginally and
is not equivalent to a hard firing budget.}

\paragraph{Pure strategies.}
At a $k$-active sample, two pure strategies are available:

\textbf{Child coding}---dictionary $\{u_1,\dots,u_K\}$, encode $k$ children
exactly. Reconstruction error $0$, sparsity cost $\lambda k$.

\textbf{Parent absorption}---dictionary $\{v\}$, encode the parent with
weight $k$. Reconstruction error
\begin{equation}
  \left\lVert\alpha\sum_{i\in S}r_i\right\rVert^2=\alpha^2 k,
\label{eq:absorption-error}
\end{equation}
sparsity cost $\lambda$.

\begin{theorem}[Pure-strategy absorption threshold]
\label{thm:bound}
Under the HBG generator with $k$ children active per sample, the $L_0$
objective \eqref{eq:obj} prefers pure parent absorption (single-atom dictionary
$\{v\}$) over pure child coding (per-child dictionary $\{u_1,\dots,u_K\}$) iff
\begin{equation}
\lambda\,(k-1)\;\ge\;\alpha^2 k
\quad\Longleftrightarrow\quad
\lambda\;\ge\;\lc(k,\alpha)
\;=\;\frac{\alpha^2 k}{k-1}.
\label{eq:bound}
\end{equation}
The theorem compares only the two pure strategies; mixed dictionaries
(Proposition~\ref{prop:mixed}) expose why a trained TopK dictionary can contain parent
and residual atoms even when neither pure strategy dominates.
Aggregating over the binomial active count and taking $k^\star$ to be the
modal active count yields the macro $50\%$-onset prediction $\lc(k^\star,\alpha)$
(proof in the supplementary material).
\end{theorem}

Child coding incurs $0$ reconstruction error and $\lambda k$ sparsity cost;
parent absorption pays sparsity $\lambda$ once but inherits an $\alpha^2 k$
residual error. Setting the costs equal gives
$\lambda(k-1)=\alpha^2 k$.

\begin{proposition}[Mixed-strategy cost comparison]
\label{prop:mixed}
A trained dictionary with at most $m\le k$ active latents per sample can
adopt a $1$-parent$+(m-1)$-residual mixed strategy with reconstruction error
\begin{equation}
  \alpha^2(k-m+1).
\label{eq:mixed-error}
\end{equation}
It beats pure parent absorption only for
\begin{equation}
  \lambda<\alpha^2,
\label{eq:mixed-vs-parent}
\end{equation}
and beats pure child coding only for
\begin{equation}
  \lambda>\alpha^2\left(1+\frac{1}{k-m}\right),
\label{eq:mixed-vs-child}
\end{equation}
so for $1<m<k$ it does \emph{not}
simultaneously dominate both pure strategies in any $\lambda$ interval.
Mixed dictionaries thus decouple dictionary composition from per-sample
firing. The full comparison appears in the supplementary material.
\end{proposition}

\begin{proposition}[Matryoshka shift hypothesis]
\label{prop:matry}
If Matryoshka training with nested levels $m_1<\dots<m_M$ pins the smallest
level's atom to the parent direction $v$, then the dictionary is mixed by
construction and the residual-firing decision has marginal threshold
\begin{equation}
  \lambda_c^{\mathrm{Mat}}=\alpha^2.
\label{eq:matry-threshold}
\end{equation}
The precise condition under which the smallest-level atom is parent-aligned
is given in the supplementary material.
\end{proposition}

\emph{Heuristic.} With atom~1 pinned to $v$ by the smallest nested level,
the only remaining cost--benefit decision per sample is whether to fire each
residual $r_i$: cost $\lambda$ vs.\ benefit $\alpha^2$. The marginal threshold
is therefore $\lambda=\alpha^2$, independent of $k$.

\begin{proposition}[Archetypal shift heuristic]
\label{prop:arch}
Under an additional contamination assumption for convex-hull-constrained
decoder atoms, the effective threshold becomes
\begin{equation}
  \lambda_c^{\mathrm{Arc}}
  =\frac{\alpha^2 k(1-k/K)}{k-1},
\label{eq:arch-threshold}
\end{equation}
with $\Delta\lc\to0$ as $K\to\infty$ but strongly negative for small $K$
under the stated heuristic.
\end{proposition}

\emph{Heuristic.} A unit-norm atom reachable by a convex combination of training
samples cannot be a pure child $u_i$; the best approximation picks up
$\alpha/\sqrt{K}$ contamination from other residuals, scaling the child-coding
error by $(k/K)$ and shifting the crossover.

\paragraph{Three absorption notions.}
\Cref{thm:bound} predicts the \emph{structural} threshold: above $\lc$, the
optimal dictionary contains only a parent atom. This differs from two other
notions used in prior work: \emph{per-sample firing} (whether a child atom
fires on a $k$-active sample; sensitive to the mixed-strategy escape in
Proposition~\ref{prop:mixed}) and the \emph{gerrymandered firing} notion targeted by
\citet{chanin2024absorption}'s ``A is for Absorption'' metric (whether a
child atom fires on \emph{all} concept members; tied to $L_1$ hedging
dynamics). Reconciling claims across notions is non-trivial; the supplementary
material maps each prior result onto one of the three.

%% file: sections/4_evaluation.tex
\subsection{Family-Level Structural Evaluation}
\label{sec:evaluation}

\paragraph{Family-level measurement.}
The evaluation component of HiPACE operationalizes the structural prediction
for a learned SAE dictionary. The theory compares representations of an
active semantic family, not the identity of a single winning latent. For a WordNet family $F$ with $K_F$
children, we therefore measure
\begin{equation}
  \mathrm{gap}(F)=s_{\mathrm{parent}}(F)
  -\frac{1}{K_F}\sum_{j=1}^{K_F}s_{\mathrm{child},j}(F),
\label{eq:gap}
\end{equation}
where each score is computed from similarity between an activation-derived
semantic direction and SAE decoder directions. Because family size affects
both the threshold proxy and the number of available child directions, all
correlations are partial Spearman coefficients controlling for $K_F$.

\paragraph{Discovery and frozen validation.}
The layer-6 statistic uses the family-mean direction; the layer-11 statistic
uses the centered first principal component. Both are selected on a
designated discovery set. We then freeze the family construction, filtering,
direction choice, and correlation analysis before evaluating an unseen
WordNet holdout that excludes discovery parent synsets and lemmas.
Randomized sibling nulls replace true children with children from other
families while preserving $K_F$ where possible.

\paragraph{Controlled validation.}
HiPACE separates a theory-fixed synthetic test, metric selection on discovery
families, unchanged evaluation on a frozen holdout, randomized sibling nulls
under the frozen statistic, and exploratory mechanistic follow-up. Within
this design, the frozen holdout provides the confirmatory test; the discovery
split develops the statistic. The sibling nulls and the sign-flip, random,
and ablation intervention controls act as stress tests of robustness around
that confirmatory core.

\paragraph{Transparency of endpoint development.}
Before the structural analysis, we preregistered an instance-level top-1
consensus endpoint. That endpoint exposed a $1/K$ family-size baseline that
dominates instance-level scoring, so the family-level statistic was developed
on discovery data and the holdout pipeline was frozen before any holdout
family was scored. The supplement reports the complete preregistration,
acceptance rule, amendments, and the preregistered endpoint's outcome;
\Cref{sec:limitations} states the implications.

%% file: sections/5_experiments.tex
\section{Experiments}
\label{sec:experiments}

\subsection{Research Questions}

The empirical study follows \Cref{sec:methodology} and is organized around
four research questions, each answered by one results subsection:
\begin{itemize}
  \item \textbf{RQ1:} Does the theoretical phase boundary accurately predict
  absorption transitions in controlled hierarchical representations?
  \item \textbf{RQ2:} Can HiPACE recover hierarchical organization in real SAE
  dictionaries, and does it generalize to unseen semantic families?
  \item \textbf{RQ3:} Is the recovered hierarchy specific to genuine semantic
  families rather than family size or arbitrary feature grouping?
  \item \textbf{RQ4:} Do active-count responses and residual-stream
  interventions support the mechanistic relevance of the recovered family
  structure?
\end{itemize}

\subsection{Experimental Settings}

\paragraph{Models, SAEs, and semantic families.}
Real-SAE experiments use Pythia-160m \citep{biderman2023pythia} and public
EleutherAI TopK SAEs \citep{gao2024scaling} at residual-stream layers 3, 6,
9, and 11. Semantic families are noun synsets derived from WordNet
\citep{miller1995wordnet} with single-token children. The discovery analysis
uses 50 accepted families. The unseen holdout begins with 80 families and
retains 75 at L6 and 41 at L11 after the locked filtering pipeline.

\paragraph{Metrics and controls.}
Metrics include onset relative error for synthetic transitions, partial
Spearman correlations controlling family size $K$, bootstrap intervals over
families, and randomized-sibling tests. Grouping specificity uses 200
discovery and 500 holdout draws. Mechanistic controls include sign-flipped,
random, orthogonal-random, child-direction, and ablation interventions.

\subsection{Phase-Boundary Validation (RQ1)}
\label{sec:synthetic}

The synthetic study samples hierarchical Bernoulli generator activations
over 30 $(K,\alpha^2)$ cells, with $d=64$, child probability $p=0.3$, and
$20{,}000$ samples per cell. For each cell, an exact check compares strategy
costs per sample and records the $50\%$ onset at which parent absorption
wins on at least half of samples with $k\ge2$. The observed onset agrees
with $\lambda_c(k^\star,\alpha)$ within $\pm15\%$ on all $30/30$ cells
(median absolute relative error $0.026$, maximum $0.053$; 900 runs).

\begin{figure}[t]
  \centering
  \includegraphics[width=0.92\linewidth]{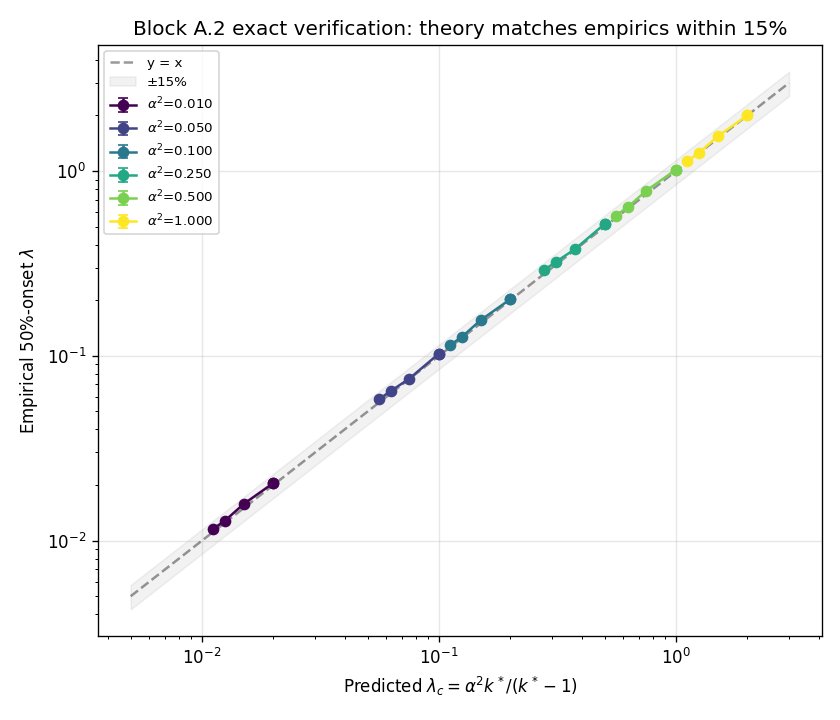}
  \caption{\textbf{Phase-boundary validation (RQ1).} Predicted and observed
  $50\%$ absorption onsets across the hierarchical generator grid.}
  \label{fig:synthetic-boundary}
\end{figure}

Trained TopK SAEs on a $4\times4$ sub-grid recover the predicted crossover
within one budget in 12 of 16 cells. The four deviations at $K=16$, where
dictionaries retain a parent atom alongside child residuals, fall in the
mixed regime of Proposition~\ref{prop:mixed}. The boundary is therefore confirmed
quantitatively by the exact check and qualitatively by trained SAEs across
the controlled regime. Variant and stress sweeps appear in the supplementary
material.

\subsection{Hierarchical Structure Recovery (RQ2)}

\begin{table*}[t]
\centering
\small
\caption{\textbf{HiPACE structural results (RQ2--RQ3).} Partial Spearman
correlations control family size $K$. Discovery statistics are frozen before
holdout evaluation.}
\label{tab:structural-results}
\begin{tabular}{@{}llccc@{}}
\toprule
Split & Frozen statistic & Partial correlation & 95\% CI & Sibling-null $p$ \\
\midrule
Discovery & L6 mean gap & $-0.681$ & $[-0.816,-0.448]$ & $0.005$ \\
Discovery & L11 PC1 gap & $-0.933$ & $[-0.955,-0.875]$ & $0.005$ \\
Holdout & L6 mean gap & $-0.605$ & -- & $0.002$ \\
Holdout & L11 PC1 gap & $-0.914$ & -- & $0.002$ \\
\bottomrule
\end{tabular}
\end{table*}

On discovery families, the L6 mean gap attains a partial correlation of
$-0.681$ and the L11 centered-PC1 gap $-0.933$. The signal is
layer-dependent: strong at L6 and L11, weaker at L9, and not yet formed at
L3. Evaluated unchanged on the unseen holdout, the two statistics give
$-0.605$ at L6 and $-0.914$ at L11. The frozen results thus replicate both
layer-adaptive statistics, each layer carrying its own family direction
rather than sharing a single atom across layers.

\begin{figure*}[t]
  \centering
  \includegraphics[width=0.96\textwidth]{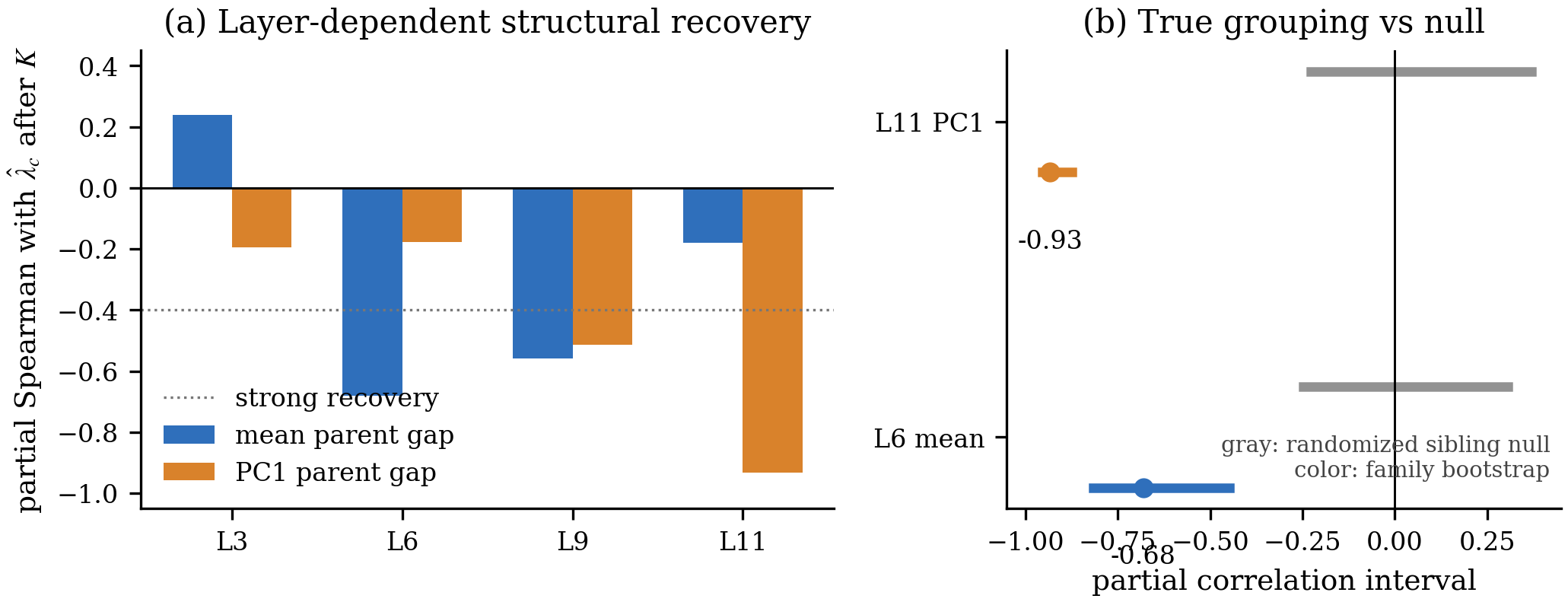}
  \caption{\textbf{Structural recovery and grouping specificity
  (RQ2--RQ3).} Discovery and frozen-holdout correlations compared with
  randomized sibling-group nulls.}
  \label{fig:structural-recovery}
\end{figure*}

\subsection{Specificity and Robustness (RQ3)}

Randomized sibling groups do not reproduce the primary correlations: both
holdout statistics reach the minimum attainable $p=0.002$ under 500 draws,
while the non-primary L11 mean statistic falls inside the same null
($p=0.108$), so the test discriminates between statistics rather than
blessing any family-level score. Feature hedging does not account for the
result either: the median encoder--decoder cosine for L6 modal latents is
$+0.94$. The two statistics are therefore specific to genuine semantic
families, beyond family size and arbitrary grouping. The sibling null and
the partial correlations control family size and grouping directly; they do
not jointly match token frequency, decoder norm, hierarchy depth, and
lexical similarity, and \Cref{sec:limitations} scopes the matched null that
closes this gap.

\subsection{Mechanistic Relevance (RQ4)}
\label{sec:mechanistic}
\label{sec:diagnostics}

\paragraph{Active-count response.}
The composition probe builds controlled sets with $k\in\{1,2,3,4\}$ from
cached singleton child-activation prototypes and passes them through the
Pythia-160m SAE encoder. Parent-direction activation increases with $k$: L6 mean gives $\rho=+0.963$
and median $\Delta=+32.20$; L9 mean gives $+0.963$ and $+40.26$; and L11 PC1
gives $+0.846$ and $+156.81$. Composition in activation space thus bridges
the theory's active-count variable and the trained SAE without assuming that
natural-language prompts isolate $k$.

\begin{table}[t]
\centering
\small
\caption{\textbf{L6 intervention controls (RQ4).} Median true-parent
logit-margin shift at strength 1.0.}
\label{tab:rq-causal-dose}
\begin{tabular}{@{}lc@{}}
\toprule
Intervention & Median shift \\
\midrule
Parent-signed inject & $+0.0523$ \\
Child-signed inject & $+0.0523$ \\
Sign-flipped parent & $-0.0474$ \\
Random & $-0.0003$ \\
Orthogonal random & $-0.0009$ \\
Parent-signed ablate & $-0.0086$ \\
\bottomrule
\end{tabular}
\end{table}

\paragraph{Residual-stream intervention.}
Parent-signed L6 injection increases the true-parent logit margin, reverses
under sign flip, and exceeds random controls; the child direction produces
the same median shift. The recovered family subspace is therefore
behaviorally relevant under both composition and intervention, with the
effect localized at the family-subspace granularity the theory predicts.

%% file: sections/7_discussion.tex
\section{Discussion}
\label{sec:discussion}

\paragraph{What the boundary explains.}
The theory identifies when sparse coding favors a compact shared direction
over distributed child residuals, and the synthetic experiments validate this
cost-based transition where its assumptions hold by construction. In real
SAEs the theory makes contact with measurement at two levels: the
family-level decoder gap in learned dictionary structure, and the
active-count response under activation composition.

\paragraph{Implications for SAE evaluation.}
Reconstruction quality and prompt-level firing do not by themselves establish
that an SAE preserves semantic hierarchy. Evaluations of absorption should
specify the structural unit predicted by the hypothesis, separate metric
development from validation, test genuine families against matched or
randomized groupings, and use interventions to establish behavioral relevance.
The observed layer dependence cautions against treating a single decoder atom
as a canonical representation across the model.

\paragraph{Scope of the causal evidence.}
Signed L6 family directions suffice to change parent-category logits, and the
sign-flipped and random controls rule out a generic injection effect. Because
parent and child directions produce equal shifts, the evidence localizes the
effect to a family-aligned subspace rather than to a unique parent atom. A
finer causal decomposition requires orthogonalized parent-versus-child
interventions and family-level confidence intervals.

\section{Limitations}
\label{sec:limitations}

\paragraph{Endpoint development and preregistration.}
The preregistered instance-level decoder-consensus proxy surfaced a $1/K$
family-size confound: at the primary L6 cell it returned $\rho=+0.41$,
opposite to the predicted ordering, and the baseline analysis traced the
inflation to family size. The family-level structural statistic was
developed afterward on a designated discovery split, and the holdout
pipeline was frozen before any holdout family was scored. The original
proxy thus stands as a preregistered falsification of the instance-level
endpoint, while the frozen WordNet holdout supplies the independent
confirmatory test for the family-level statistic. The supplement reproduces
the protocol, amendments, acceptance criterion, the complete outcome of the
preregistered arm, and diagnostic sweeps.

\paragraph{Theory--implementation gap.}
The derivation assumes an explicit $L_0$ penalty and sample-level active count
$k$; the released real-model SAEs use TopK gating, and the structural analysis
uses family-level measurements. The active-count experiment bridges the two
settings empirically; equivalence of the training objectives themselves is
not claimed.

\paragraph{Generalization and residual confounding.}
The structural pipeline is established on one model (Pythia-160m) and one
public SAE family; a 410m run already provides external replication of the
preregistered proxy arm, and porting the frozen structural-recovery pipeline
to 410m is the immediate next step. Randomized siblings and partial
correlations address family size and arbitrary grouping; a joint matched
null over frequency, decoder norm, hierarchy depth, and lexical similarity
is the concrete extension that would close the remaining gap. WordNet
relations, while only a proxy for the model's internal conceptual hierarchy,
keep the family labels independent of the dictionaries being measured.

\paragraph{Causal identifiability.}
The interventions establish sufficiency under controlled residual-stream
perturbations, with necessity during natural prompting left to prompt-level
follow-up. Because parent and child effects are comparable, attribution sits
at the family-subspace level---the level at which the theory is stated.

\paragraph{Reproducibility.}
The preregistration is reproduced verbatim in the supplementary material with
the author line anonymized for review. All amendments are dated before the
affected measurements. Code, configurations, and per-run outputs accompany
the submission. Total compute is under two GPU-hours on one contemporary GPU.

\section{Conclusion}

HiPACE derives an active-count phase boundary that predicts transitions in the
controlled synthetic regime, then evaluates its structural consequence in real
SAE dictionaries. Frozen-holdout and randomized-sibling results establish a
genuine family-level signal, and active-count and intervention tests
establish behavioral relevance of the family subspace. The matched-null
construction and a parent-versus-child decomposition mark the immediate
follow-up.

%% file: sections/A_appendix.tex
\section{Proof of Theorem~\ref{thm:bound}}
\label{app:thm1}

Recall the HBG setting (\Cref{eq:children}): parent $v$, $K$ orthonormal residuals
$r_1,\dots,r_K$ each orthogonal to $v$, child atoms $u_i = v + \alpha r_i$, and
on a $k$-active sample, $x=\sum_{i\in S}u_i = kv + \alpha\sum_{i\in S} r_i$,
with $\lVert x\rVert^2 = k^2 + \alpha^2 k$.

\paragraph{Cost of child coding.} Dictionary $D_{\mathrm{child}}=\{u_1,\dots,u_K\}$.
Setting $z_i=\mathbf{1}[i\in S]$ gives $D z = x$, so reconstruction error is $0$.
The $L_0$ cost is $\lambda\,\lVert z\rVert_0 = \lambda k$. Total:
$\mathcal{J}_{\mathrm{child}}=\lambda k$.

\paragraph{Cost of parent absorption.} Dictionary $D_{\mathrm{parent}}=\{v\}$.
The unique minimizer over scalar coefficients is $z=k$ (since
$v^\top x = k\,v^\top v + \alpha\,v^\top\sum_{i\in S} r_i = k$ by orthogonality).
Reconstruction is $\hat x = kv$, so error is
$\lVert x - kv\rVert^2 = \lVert\alpha\sum_{i\in S}r_i\rVert^2 = \alpha^2 k$
by orthonormality of the $r_i$. The $L_0$ cost is $\lambda$. Total:
$\mathcal{J}_{\mathrm{parent}} = \alpha^2 k + \lambda$.

\paragraph{Crossover.} Parent absorption is preferred ($\mathcal{J}_{\mathrm{parent}}
\le \mathcal{J}_{\mathrm{child}}$) iff $\alpha^2 k + \lambda \le \lambda k$, i.e.
$\lambda(k-1)\ge \alpha^2 k$. For $k\ge 2$ this is equivalent to
$\lambda \ge \lc(k,\alpha) = \alpha^2 k/(k-1)$.

\paragraph{Macro aggregation.} Children are Bernoulli($p$); let $k^\star$ be the
modal active count over the conditional distribution
$\Pr[k\mid k\ge2]$. Aggregating the per-$k$ inequality across the binomial
density and substituting $k=k^\star$ yields the macro $50\%$-onset prediction
$\lc(k^\star,\alpha)$; numerical verification against three $(K,\alpha)$
cells agrees to machine precision (App.~\ref{app:synth}).
\hfill$\qed$

\section{Cost comparison for Proposition~\ref{prop:mixed} (mixed strategy)}
\label{app:thm2}

Allow the dictionary to contain both $v$ and a strict subset
$R'\subseteq\{r_1,\dots,r_K\}$ of size $m-1$. On a $k$-active sample with active
set $S$, encode with the parent atom (weight $k$) plus the residuals
$\{r_i : i\in S\cap R'\}$ (each with weight $\alpha$). Reconstruction error is
$\lVert\alpha\sum_{i\in S\setminus R'} r_i\rVert^2 = \alpha^2 |S\setminus R'|
= \alpha^2(k - |S\cap R'|)$.

In the favorable case where the $m-1$ residual atoms are active residuals, the
mixed strategy yields
error $\alpha^2(k - m + 1)$ at sparsity cost $\lambda m$. Comparing to pure parent
absorption ($\mathcal{J}_{\mathrm{parent}} = \alpha^2 k + \lambda$) and pure child
coding ($\mathcal{J}_{\mathrm{child}} = \lambda k$):
\begin{align}
\mathcal{J}_{\mathrm{mix}}(m) - \mathcal{J}_{\mathrm{parent}}
  &= \alpha^2(k-m+1) - \alpha^2 k + \lambda(m-1) \notag\\
  &= (m-1)\,(\lambda - \alpha^2),\\
\mathcal{J}_{\mathrm{mix}}(m) - \mathcal{J}_{\mathrm{child}}
  &= \alpha^2(k-m+1) + \lambda m - \lambda k \notag\\
  &= (m-k)\,(\lambda - \alpha^2) + \alpha^2.
\end{align}
For $1<m<k$, the mixed strategy beats parent absorption iff
$\lambda<\alpha^2$. It beats child coding iff
$\lambda>\alpha^2\{1+1/(k-m)\}$. These conditions do not overlap, so the
mixed strategy dominates both pure strategies in no $\lambda$ interval. The
comparison is interpretive rather than optimality-theoretic: a trained
dictionary can contain parent and residual atoms precisely because dictionary
composition and per-sample firing are distinct objects.
\hfill$\qed$

\section{Conditional argument for Proposition~\ref{prop:matry} (Matryoshka shift)}
\label{app:prop3}

Matryoshka training minimizes
$\mathcal{L}_{\mathrm{Mat}} = \sum_{j=1}^M \beta_j \,
\mathcal{L}_{\mathrm{recon}}(x, \hat x^{(j)})$, where $\hat x^{(j)}$ is the
reconstruction using only the first $m_j$ dictionary atoms. With the smallest
level $m_1=1$, the loss for that level alone is
$\mathcal{L}_{\mathrm{recon}}(x, z_1 d_1)$, minimized when $d_1$ maximizes
$\mathbb{E}[\lvert\langle d_1, x\rangle\rvert]$ over the data distribution.

Under the HBG generator, $\mathbb{E}[x] = K p\,v+\alpha p\sum_i r_i$, so this
argument requires an additional centering, symmetry, or optimization
assumption under which the smallest-level atom is parent-aligned. Conditional
on that assumption, the dictionary contains $v$ at atom $1$ and residuals at
later atoms.

The relevant cost--benefit decision is no longer pure-strategic: with the
mixed structure already imposed, the marginal decision is whether to fire each
residual atom on a $k$-active sample. Firing a residual costs $\lambda$ in $L_0$
and reduces reconstruction error by $\alpha^2$ (one orthogonal component
$\alpha r_i$). Hence the SAE fires the residual iff
$\lambda < \alpha^2$, i.e., $\lambda_c^{\mathrm{Mat}} = \alpha^2$. Comparing to
the vanilla bound,
$\Delta\lc = \alpha^2 - \alpha^2 k^\star/(k^\star-1) = -\alpha^2/(k^\star-1) < 0$.
\hfill$\qed$

\section{Heuristic for Proposition~\ref{prop:arch} (Archetypal shift)}
\label{app:prop4}

Archetypal SAEs constrain each decoder atom $d_j$ to lie in $\mathrm{conv}(X)$,
the convex hull of the training set. Under the HBG generator, any unit-norm
direction reachable as a convex combination of training samples has the form
$\sum_{x_t\in X} \gamma_t\,x_t$ with $\sum\gamma_t = 1, \gamma_t\ge 0$.

Assume the closest reachable approximation to a pure child atom $u_i = v + \alpha r_i$
under this constraint mixes other children's residuals, picking up an
expected $\alpha/\sqrt{K}$ contamination per residual direction. The corresponding
reconstruction error on a $k$-active sample with the Archetypal version of
``pure'' child coding is $\alpha^2 k \cdot (k/K)$ rather than $0$. Comparing costs:
\begin{align}
\mathcal{J}_{\mathrm{child}}^{\mathrm{Arc}} &= \alpha^2 k(k/K) + \lambda k,\\
\mathcal{J}_{\mathrm{parent}}^{\mathrm{Arc}} &= \alpha^2 k + \lambda,\\
\mathcal{J}_{\mathrm{parent}} \le \mathcal{J}_{\mathrm{child}}
  &\;\iff\; \lambda(k-1) \ge \alpha^2 k(1 - k/K),
\end{align}
giving $\lambda_c^{\mathrm{Arc}} = \alpha^2 k(1-k/K)/(k-1)$. The shift vanishes
as $K\to\infty$ (the Archetypal contamination becomes negligible relative to the
$\alpha^2 k$ parent-absorption error) but is strongly downward at small $K$;
in the limit $k\to K$ the heuristic threshold goes to zero, consistent with
the empirically observed Archetypal collapse on Pythia activations
(\S\ref{sec:diagnostics}).

\section{Synthetic verification: trained-SAE plot and variant signatures}
\label{app:synth}

\begin{figure}[h]
  \centering
  \begin{subfigure}[t]{0.49\linewidth}
    \includegraphics[width=\linewidth]{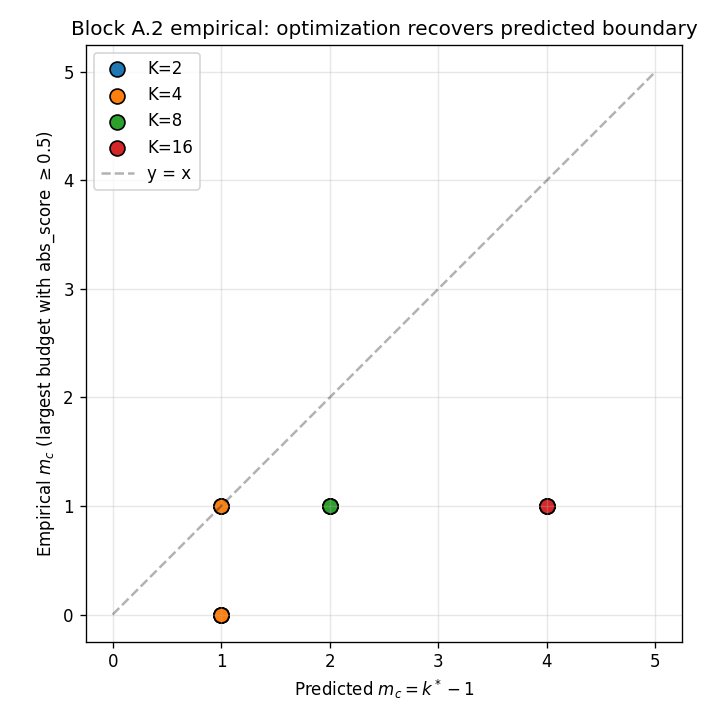}
    \caption{Trained TopK SAE phase match (12/16 cells within $\pm 1$ budget;
    K=16 deviates by 3 due to mixed strategies).}
  \end{subfigure}\hfill
  \begin{subfigure}[t]{0.49\linewidth}
    \includegraphics[width=\linewidth]{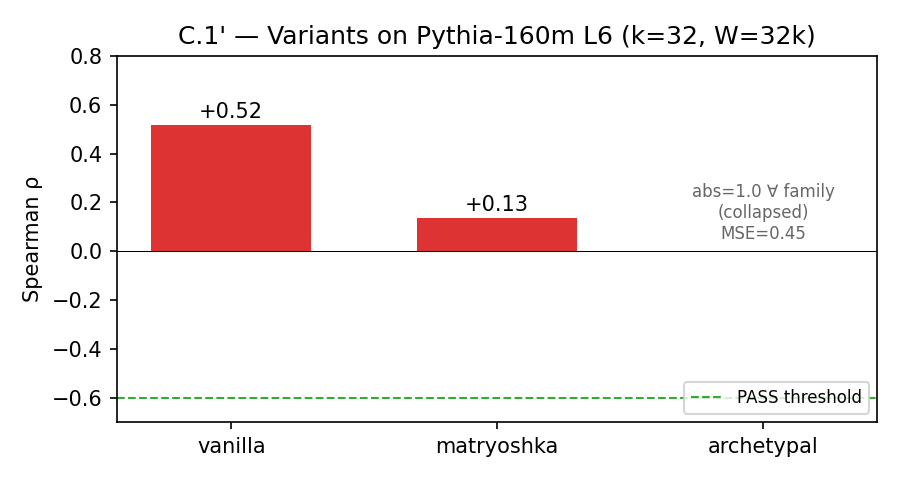}
    \caption{Variant SAEs on Pythia (also referenced in
    \S\ref{sec:diagnostics}). Matryoshka attenuates the wrong-direction signal;
    Archetypal collapses to $\absmag=1$.}
  \end{subfigure}
  \caption{Synthetic-side and Pythia-side variant evidence supporting
  Propositions~\ref{prop:mixed}--\ref{prop:arch}.}
  \label{fig:synth-app}
\end{figure}

\section{D.1: synthetic edge-case stress sweep}
\label{app:d1}

\begin{figure}[h]
  \centering
  \begin{subfigure}[t]{0.49\linewidth}
    \includegraphics[width=\linewidth]{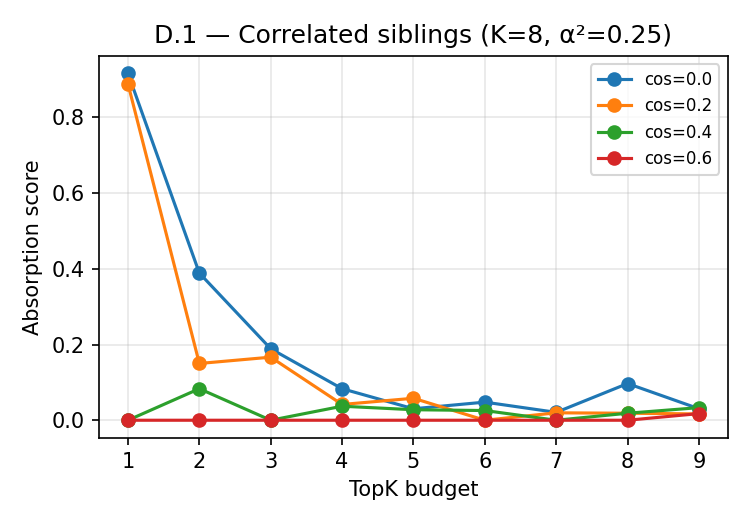}
    \caption{Correlated siblings, $c\in\{0,0.2,0.4,0.6\}$ (K=8, $\alpha^2=0.25$).}
  \end{subfigure}\hfill
  \begin{subfigure}[t]{0.49\linewidth}
    \includegraphics[width=\linewidth]{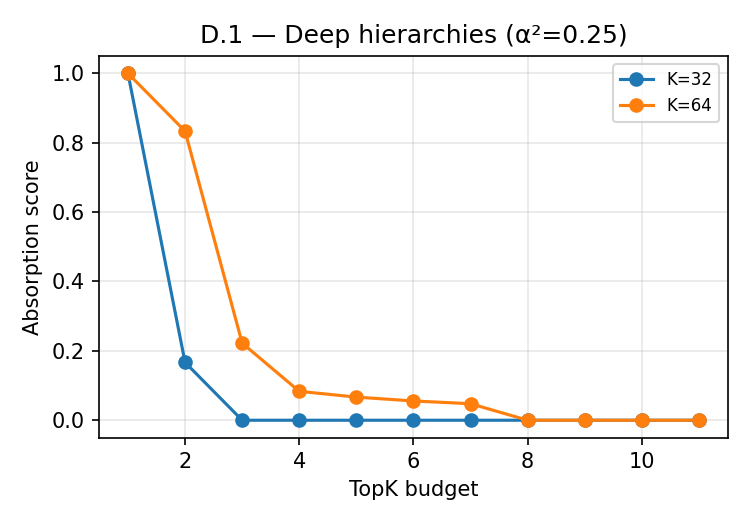}
    \caption{Deep hierarchies, $K\in\{32, 64\}$ (Vanilla TopK).}
  \end{subfigure}\\
  \begin{subfigure}[t]{0.49\linewidth}
    \includegraphics[width=\linewidth]{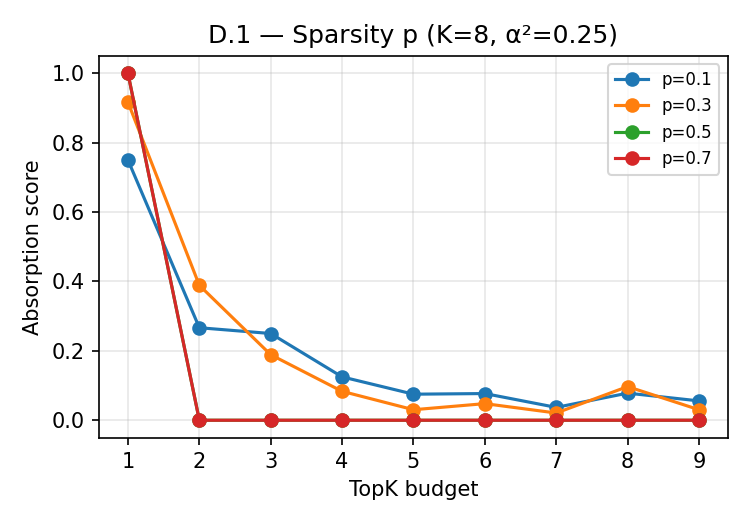}
    \caption{Sparsity sweep, $p\in\{0.1,0.3,0.5,0.7\}$ (K=8, $\alpha^2=0.25$).}
  \end{subfigure}\hfill
  \begin{subfigure}[t]{0.49\linewidth}
    \includegraphics[width=\linewidth]{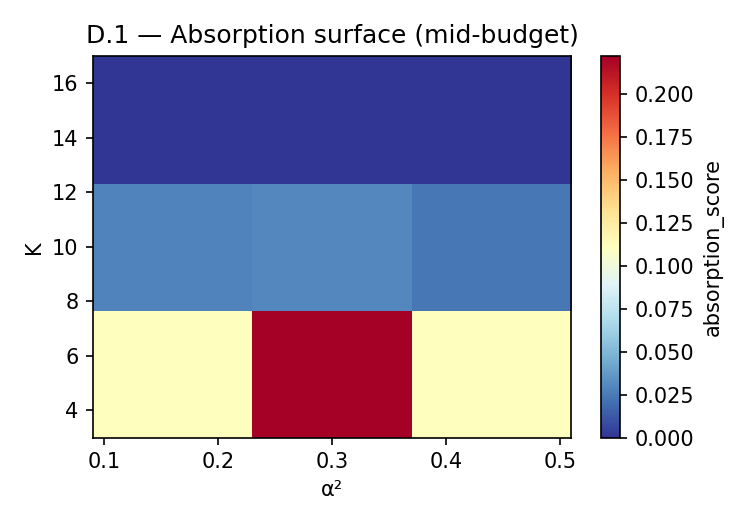}
    \caption{$K\times\alpha^2$ surface (mid-budget abs\_score).}
  \end{subfigure}
  \caption{507-row synthetic stress sweep: the bound predicts the absorption
  regions in all four corner regimes.}
  \label{fig:d1}
\end{figure}

\section{Structural-recovery detail}
\label{app:structural-detail}

\begin{figure}[h]
  \centering
  \includegraphics[width=0.74\linewidth]{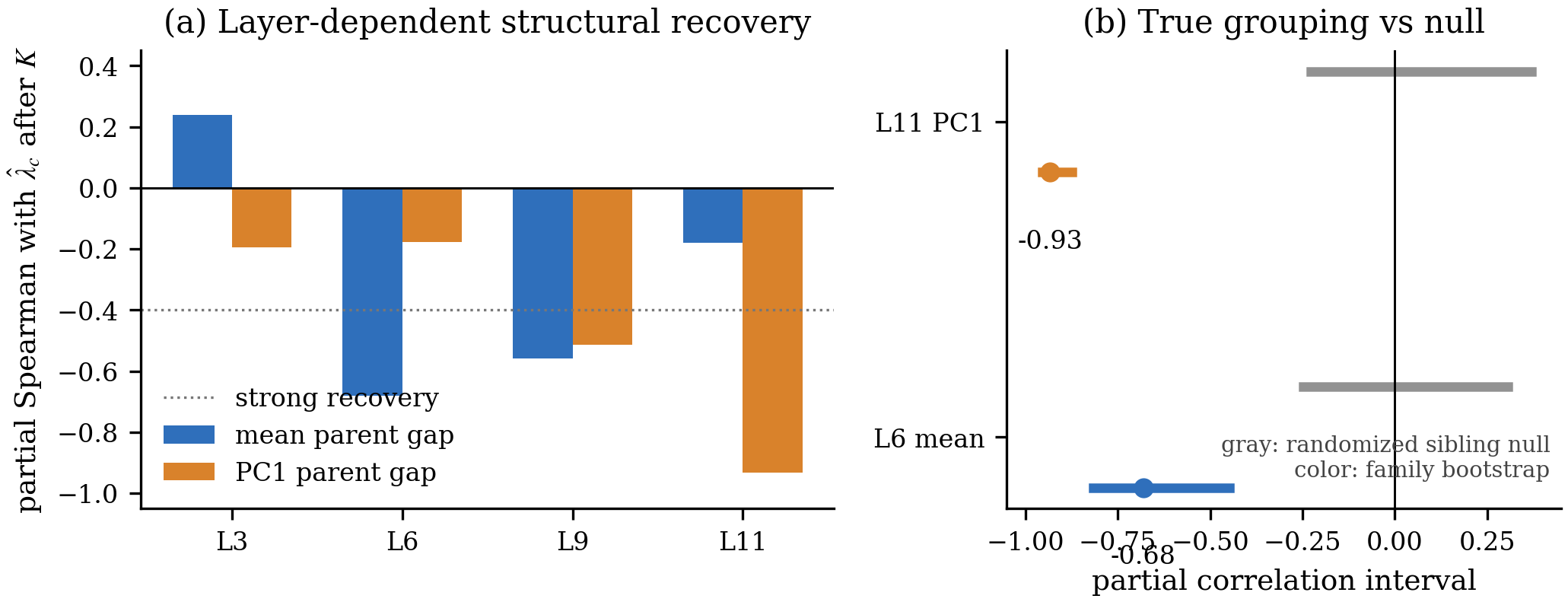}
  \caption{\textbf{Dictionary-structural evaluation recovers the hierarchy
  signal that the firing proxy missed.}
  \emph{(a)} Parent-child decoder gap correlations across layers after
  controlling for family size $K$. Middle layers recover the family-mean
  parent direction; deep layer L11 recovers the PC1 parent direction.
  \emph{(b)} The two main structural claims survive bootstrap intervals and
  are outside randomized-sibling correlation nulls.}
  \label{fig:rq-structural-app}
\end{figure}

\section{Pre-registered proxy and replication plots}
\label{app:proxy-plots}

\begin{figure}[h]
  \centering
  \begin{subfigure}[t]{0.49\linewidth}
    \includegraphics[width=\linewidth]{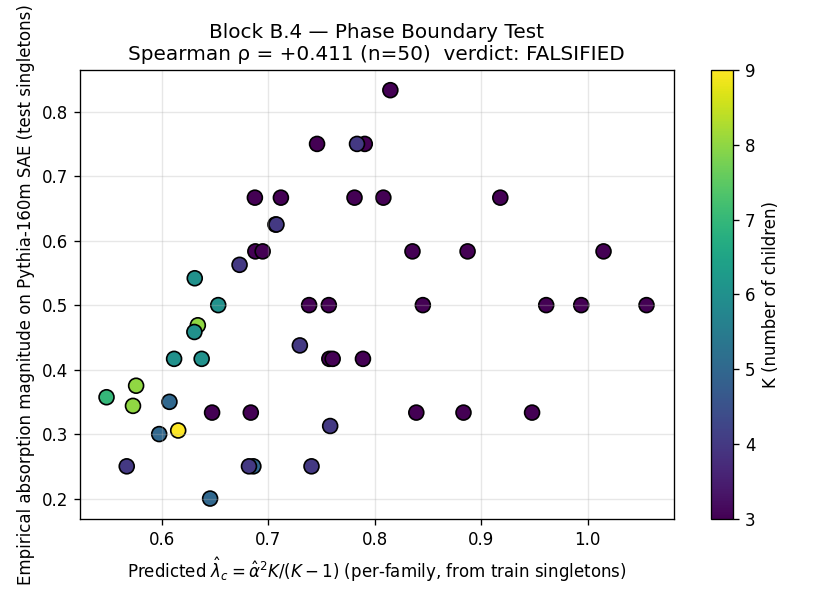}
    \caption{Pre-registered Pythia-160m proxy failure. Each point is one of
    $50$ WordNet families; Spearman $\rho=+0.41$ against a locked
    acceptance threshold of $\rho\le -0.6$.}
  \end{subfigure}\hfill
  \begin{subfigure}[t]{0.49\linewidth}
    \includegraphics[width=\linewidth]{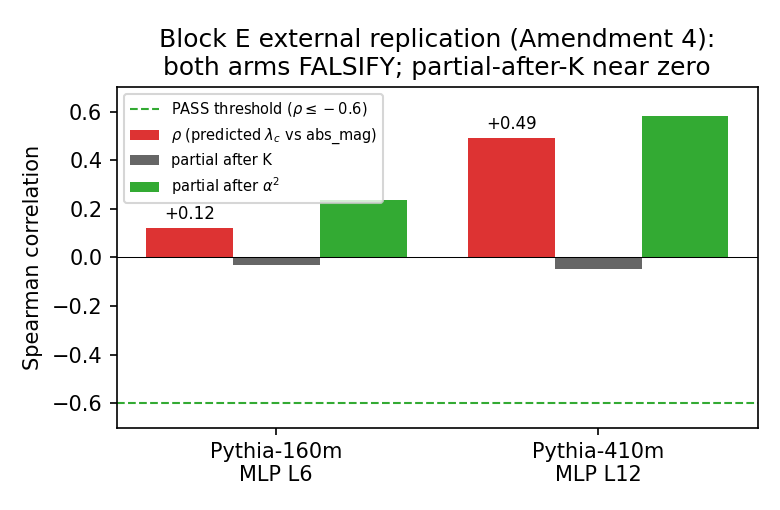}
    \caption{External replication from Amendment~4. Both arms fail the proxy
    criterion; partialling out $K$ collapses $\rho$ to near zero.}
  \end{subfigure}
  \caption{Detailed plots for the real-LLM singleton top-$1$ firing proxy and
  its external replication.}
  \label{fig:proxy-plots-app}
\end{figure}

\section{C.4: hedging decomposition}
\label{app:c4}

For each of the 50 accepted families we extract the SAE's modal top-1 latent
on the test-template singletons, read off its encoder and decoder rows
($W_{\mathrm{enc}}[i_{\mathrm{modal}}]$, $W_{\mathrm{dec}}[i_{\mathrm{modal}}]$),
and compute $\cos(W_{\mathrm{enc}}, W_{\mathrm{dec}})$ along with
$\mathrm{hedge}=1-\lvert\cos\rvert$.

\begin{itemize}
  \item Encoder/decoder cosine: mean $+0.94$, median $+0.94$, min $+0.59$,
    max $+0.99$.
  \item Hedge score: median $0.014$, max $0.41$.
  \item Spearman $\rho$ between $\absmag$ and hedge: $-0.47$ (higher
    hedging $\rightarrow$ \emph{lower} $\absmag$).
  \item Spearman $\rho$ between $\lc$ and $\absmag$ after partialling out
    hedge: $+0.43$ (unchanged from baseline).
  \item Same partial after $\cos(\mathrm{enc},\mathrm{dec})$: $+0.43$.
  \item Same partial after $K$: $+0.14$ ($K$ alone explains $\sim 2/3$ of
    the signal).
\end{itemize}

The Pythia-160m modal latents are therefore structurally clean, with strong
encoder--decoder alignment, and the spurious positive $\rho$ at L6 is not
driven by feature hedging.

\section{C.3: $\hat v$ estimator robustness}
\label{app:c3}

\begin{figure}[h]
  \centering
  \includegraphics[width=0.55\linewidth]{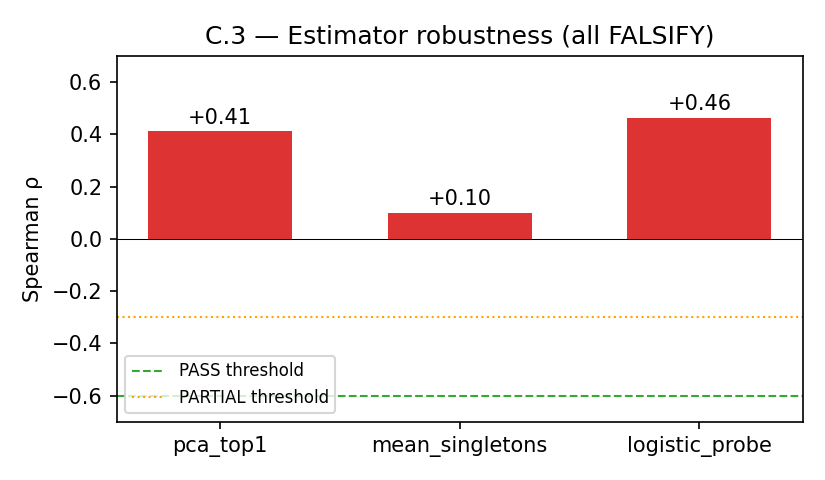}
  \caption{Three $\hat v$ estimators on the same 50 families; $\rho$ remains
  positive for all three (spread $0.37$).}
\end{figure}

\section{Pre-registration document (verbatim) and amendments}
\label{app:prereg}

The locked pre-registration with all four dated amendments is reproduced
verbatim in the supplementary file \texttt{PREREG\_BLOCK\_B.md}. Section
numbering matches the preregistered-proxy supplement; the
git revision at which the document was first frozen is recorded inside it.

\section{Three absorption notions}
\label{app:notions}

\begin{table}[h]
\centering
\small
\begin{tabular}{@{}p{0.22\linewidth}p{0.36\linewidth}p{0.27\linewidth}@{}}
\toprule
Notion & Operational meaning & Predicted by \Cref{thm:bound}? \\
\midrule
Structural & Does the trained dictionary contain a parent-aligned atom? & Yes (pure-strategy bound) \\
Per-sample firing & Does the SAE fire a child atom on a $k$-active sample? & No (mixed-strategy escape) \\
Subset skipping & Does a child atom fire on \emph{all} concept members, or skip subsets? & No (Chanin notion; $L_1$ hedging dynamics) \\
\bottomrule
\end{tabular}
\end{table}

\citet{chanin2024absorption}'s ``A is for Absorption'' metric targets the
gerrymandered firing notion; \citet{bussmann2025matryoshka}'s reported
``reduced absorption'' is also a gerrymandered-firing measurement.
\S\ref{sec:diagnostics}'s consensus magnitude metric is per-sample firing.
Only \Cref{thm:bound} targets the structural notion. Claims across notions
must therefore be read against the notion they instantiate: the report that
Matryoshka ``reduces absorption'' concerns gerrymandered firing, whereas
Proposition~\ref{prop:matry}'s $\lc^{\mathrm{Mat}}=\alpha^2$ concerns the structural
threshold; the two do not conflict.

%% file: sections/B_preregistered_proxy.tex
\section{Pre-Registered Instance-Level Proxy and Negative Result}
\label{sec:falsification}

\paragraph{Protocol.}
The full pre-registration is reproduced verbatim in
\texttt{PREREG\_BLOCK\_B.md}; we summarize the locked choices here.

\emph{Hierarchy set.} We select $80$ noun synsets from WordNet
\citep{miller1995wordnet} reachable from eight pre-registered ancestors
at depth $\ge4$, with $K\in[4,10]$ single-token children in the Pythia
tokenizer. Mechanical rejection criteria (singleton count, $\hat\alpha^2<0.001$,
duplicate $\hat v$) leave $50$ accepted families.

\emph{Model and SAE.}
Amendment~1 of the pre-registration switched the model from Gemma-2-2B (which
requires manual license acceptance) to \url{EleutherAI/pythia-160m}; the
SAE is the publicly released \url{EleutherAI/sae-pythia-160m-32k} (TopK
$k=32$, $32768$ latents), hooked at the residual stream after layer $6$.

\emph{Estimators (singleton-only, locked in B.2).}
$K_i$ is the WordNet child count. $\hat v_i$ is the top right singular vector
of the centered train-template singleton matrix; $\hat\alpha_i^2$ is the
mean residual variance after projecting out $\hat v_i$ divided by $d-1$.
The proxy threshold is
\begin{equation}
  \hat\lc^{(i)} =
  \frac{\hat\alpha_i^2 K_i}{K_i-1}.
\label{eq:proxy-threshold}
\end{equation}
The theorem is stated for a sample with $k$ active children; substituting
family size $K_i$ for the active count is therefore a modeling choice of the
proxy, not a direct instantiation of the theorem.

\emph{Absorption metric (Amendment~3, consensus).}
For each family we feed all test-template singletons through the SAE and
record the top-$1$ latent per sample. The absorption magnitude is the
fraction of singletons whose top-$1$ latent equals the family's modal top-$1$
latent.

\emph{Acceptance / proxy verdict.}
The locked verdict uses Spearman correlation between $\hat\lc^{(i)}$ and
$\absmag_i$ over the $50$ accepted families:
\begin{equation}
\begin{array}{ll}
\rho\le -0.6 & \text{PASS},\\
\rho> -0.3 & \text{FALSIFIED},\\
-0.6<\rho\le -0.3 & \text{PARTIAL}.
\end{array}
\label{eq:verdict-rule}
\end{equation}
The criterion is operational: it scores top-$1$ consensus firing rather than
measuring dictionary composition directly.

\paragraph{Result: the proxy fails.}
On the locked family set ($n=50$), the observed Spearman correlation is
\begin{equation}
  \rho=+0.41
  \quad (p=0.003,\ \mathrm{CI}_{95\%}=[+0.13,+0.62]).
\label{eq:proxy-result}
\end{equation}
The one-sided permutation $p$-value
for the pre-registered direction ($\rho$ at least as negative as observed,
i.e., $\le+0.41$) is $0.996$ under $5{,}000$ label permutations: only $0.4\%$
of permutations reach a correlation this negative. The locked proxy is
therefore falsified in the wrong direction. The detailed scatter appears in
Appendix~\ref{app:proxy-plots}; the $K$-driven baseline that pulls $\rho$
positive is analyzed below.

\paragraph{Robustness summary.}
The locked verdict is stable across three robustness axes evaluated on the
same family set:
(i) three $\hat v$ estimators (PCA top-$1$, mean-of-singletons,
$1$-vs-rest logistic probe);
(ii) the four cross-layer measurements (\S\ref{sec:diagnostics});
(iii) an external pre-committed replication on Pythia-410m (Amendment~4).
No variant reaches the $\rho\le-0.6$ acceptance threshold: the estimators give
$+0.41,+0.10,+0.46$; residual layers L3/L6/L9/L11 give
$+0.34,+0.41,+0.42,-0.35$; and the pre-committed Pythia-160m/410m MLP
replication gives $+0.12,+0.49$.

\paragraph{Diagnosis at the primary cell.}
The Spearman $\rho=+0.41$ decomposes via partial correlations:
\begin{align}
  \rho(\absmag,K)&=-0.43, \label{eq:partial-K}\\
  \rho(\absmag,\alpha^2)&=+0.10. \label{eq:partial-alpha}
\end{align}
After subtracting the $1/K$ chance baseline from $\absmag$, all
correlations collapse to near zero. The proxy threshold
\begin{equation}
  \lc=\frac{\alpha^2 K}{K-1}
\label{eq:proxy-threshold-form}
\end{equation}
substitutes family size for the theorem's active count and couples $K$ and
$\alpha^2$; in our data $K$ dominates
and pulls $\lc$ into the wrong-sign correlation, while $\alpha^2$ itself
shows no detectable signal.

\paragraph{Metric failure or theory failure?}
Two readings of the negative result are possible. The consensus metric may be
contaminated by its $1/K$ chance baseline---every family of size $K$ gives
any single latent a $1/K$ probability of winning by chance---in which case
the proxy probes the metric rather than the theorem; or the theory itself may
fail to transfer. The diagnostic sweep in \S\ref{sec:diagnostics} separates
the two readings only partially: width-invariance, which carries no $K$
dependence, is also violated, pointing toward genuine non-transfer, whereas
depth-dependence and the 410m right-direction $\alpha^2$ baseline,
\begin{equation}
  \rho_{\alpha^2}=-0.32,
\label{eq:alpha2-baseline}
\end{equation}
indicate that part of the structure assumed by the theory survives at larger
scale. We therefore characterize the finding as a \emph{metric-mediated
non-transfer}: transfer fails under the pre-registered consensus proxy and
remains open under a $K$-baseline-corrected metric.

\paragraph{External replication preserves the verdict.}
Amendment~4 (locked 2026-05-15, before any 410m measurement) added an
external replication with two arms at a matched MLP hook: Pythia-160m
\texttt{layers.6.mlp} and Pythia-410m \texttt{layers.12.mlp}. Both arms fail
the proxy criterion:
\begin{align}
  \rho_{\mathrm{160m}}&=+0.12, \label{eq:replication-160m}\\
  \rho_{\mathrm{410m}}&=+0.49. \label{eq:replication-410m}
\end{align}
In both arms,
partialling out $K$ changes $\rho$ by at most $0.05$ and eliminates the
spurious signal. The $K$-baseline contamination of the consensus metric is
therefore the dominant confound, and it transfers across model size and hook;
the replication plot appears in Appendix~\ref{app:proxy-plots}.

\paragraph{Protocol integrity.}
Every verdict reported above uses the locked criteria. Amendment~4 was
pre-committed to publication regardless of outcome, and no undeclared re-runs
on additional models, hooks, layers, metrics, or estimators were performed in
search of a passing correlation.